\PassOptionsToPackage{colorlinks=true, linkcolor=red, citecolor=magenta, urlcolor=blue}{hyperref}
\PassOptionsToPackage{dvipsnames}{xcolor}
\documentclass[nonacm,sigconf]{acmart}

\AtBeginDocument{%
  }

\usepackage{tabularray}
\usepackage{soul}
\usepackage{multirow}
\usepackage{enumitem}

\hypersetup{
  colorlinks=true,
  linkcolor=red,
  citecolor=magenta,
  urlcolor=blue
}
\begin{document}


\title[DF-P2E: From Prediction to Explanation]{From Prediction to Explanation: Multimodal, Explainable, and Interactive Deepfake Detection Framework for Non-Expert Users}


\author{Shahroz Tariq}
\email{shahroz.tariq@data61.csiro.au}
\affiliation{%
  \institution{Data61, CSIRO, Australia}
  \country{}
  }\authornote{Equal contribution} \authornote{Corresponding author}

\author{Simon S. Woo}
\email{swoo@g.skku.edu}
\affiliation{%
  \institution{Sungkyunkwan University, S. Korea}
  \country{}
}

\author{Priyanka Singh}
\email{priyanka.singh@uq.edu.au}
\affiliation{%
  \institution{University of Queensland, Australia}
  \country{}
}\authornotemark[1]

\author{Irena Irmalasari}
\email{i.irmalasari@uq.edu.au}
\affiliation{%
  \institution{University of Queensland, Australia}
  \country{}
}

\author{Saakshi Gupta}
\email{saakshi.gupta@uq.edu.au}
\affiliation{%
  \institution{University of Queensland, Australia}
  \country{}
}

\author{Dev Gupta}
\email{dev.gupta@uq.edu.au}
\affiliation{%
  \institution{University of Queensland, Australia}
  \country{}
}

\renewcommand{\shortauthors}{Tariq et al.}

\begin{abstract}
The proliferation of deepfake technologies poses urgent challenges and serious risks to digital integrity, particularly within critical sectors such as forensics, journalism, and the legal system. While existing detection systems have made significant progress in classification accuracy, they typically function as black-box models, offering limited transparency and minimal support for human reasoning. This lack of interpretability hinders their usability in real-world decision-making contexts, especially for non-expert users. In this paper, we present \textbf{DF-P2E (Deepfake: Prediction to Explanation)}, a novel multimodal framework that integrates visual, semantic, and narrative layers of explanation to make deepfake detection interpretable and accessible. The framework consists of three modular components: (1) a deepfake classifier with Grad-CAM-based saliency visualisation, (2) a visual captioning module that generates natural language summaries of manipulated regions, and (3) a narrative refinement module that uses a fine-tuned Large Language Model (LLM) to produce context-aware, user-sensitive explanations. We instantiate and evaluate the framework on the DF40 benchmark, the most diverse deepfake dataset to date. Experiments demonstrate that our system achieves competitive detection performance while providing high-quality explanations aligned with Grad-CAM activations. By unifying prediction and explanation in a coherent, human-aligned pipeline, this work offers a scalable approach to interpretable deepfake detection, advancing the broader vision of trustworthy and transparent AI systems in adversarial media environments.
\end{abstract}



\begin{teaserfigure}
\centering
  \includegraphics[trim={17pt 186pt 373pt 17pt},clip,width=0.454\linewidth]{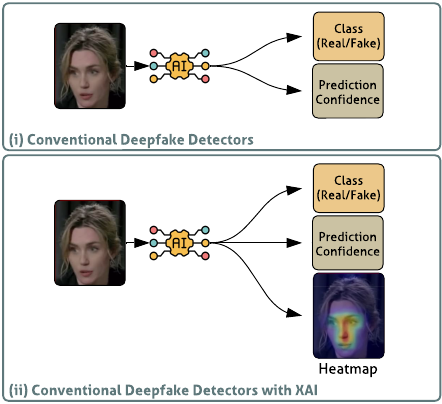}\hfill
  \includegraphics[trim={17pt 17pt 373pt 212pt},clip,width=0.52\linewidth]{figures/DF_XAI.pdf}
  \vspace{-10pt}
  \caption{Comparison of DF-P2E with existing deepfake detection pipelines.} 
  \label{fig:teaser}
\end{teaserfigure}



\maketitle
\footnotetext[1]{This paper has been accepted to the 33rd ACM International Conference on Multimedia (MM '25), October 27--31, 2025, Dublin, Ireland.}

\section{Introduction}
\label{sec:intro}

The emergence of deepfake media, synthetically generated visual or audiovisual content created using deep neural networks, has catalysed a paradigm shift in the manipulation of digital content~\cite{jafarforensics}. Once confined to academic prototypes, deepfakes have evolved into sophisticated tools that can convincingly replicate facial expressions~\cite{hosler2021deepfakes}, speech patterns, and even emotional cues of real individuals~\cite{mittal2020emotions}. Although these technologies possess legitimate potential in domains such as the entertainment industry~\cite{wipo2021deepfakes}, education~\cite{Reed2024BrightSideDeepfakes}, and accessibility~\cite{philmlee2023practice}, their misuse has outpaced regulatory and technical safeguards~\cite{Colman2025AI_Fraud_Surge,ShahrozAWS,tariq2023evaluating,WDC_Metaverse}. Incidents of political disinformation~\cite{Bond2024DeepfakesElections}, biometric identity theft~\cite{CaseIQ2024DeepfakeIdentityTheft}, reputational sabotage~\cite{hosler2021deepfakes}, and financial fraud~\cite{CNN2024DeepfakeCFOScam} have been documented worldwide, with a recent case involving the use of a deepfake video call to defraud a Hong Kong executive of \$25 million~\cite{CNN2024DeepfakeCFOScam}.

In response, the field of deepfake detection has rapidly advanced. State-of-the-art detection systems based primarily on Convolutional Neural Networks (CNNs), vision transformers, and multimodal embeddings~\cite{lifacexray, liexposingdeepfake, nicore, le2025soksystematizationbenchmarkingdeepfake,ShallowNet1,ShallowNet2,CLRNetold,tariq2021clrnet,SamTAR,SamGAN,MinhaFRETAL,MinhaCORED,HasamACMMM,JeonghoPTD}—have achieved impressive classification accuracy on large-scale datasets such as FaceForensics++~\cite{rossler2019faceforensics++}, FakeAVCeleb~\cite{khalid2021fakeavceleb} and DF40~\cite{yan2024df40}. However, despite their predictive success, most of these systems operate as \emph{black-box classifiers}. They produce binary labels or confidence scores with little or no insight into their internal reasoning~\cite{WDC_Why,cho23_RWDF}. This lack of transparency and trustworthiness critically limits their utility in high-stakes domains such as digital forensics, journalism, and court proceedings, where the provision of trust, traceability, and interpretability is of paramount importance ~\cite{chalkidisdeeplearninginlaw, lippiclaudette, blackboxlegal, explainablereasoninglegal, Tariq2024a, Tariq2024,tariq2025llms}.


Explainable Artificial Intelligence (XAI) offers a partial remedy~\cite{pinhasovadversarialdetection}. Techniques such as Grad-CAM~\cite{selvaraju2017grad}, SHAP~\cite{lundberg2017unified}, and LIME~\cite{lime} can highlight salient visual regions that influence a model's decision. However, these visual cues alone—heatmaps overlaid on faces—are often insufficient for non-expert stakeholders. Studies show that saliency maps, while useful for developers, are difficult for laypersons to interpret without contextual scaffolding~\cite{alqaraawi2020evaluating}. To build public trust and support decision-making, interpretability must go beyond visualisation: it must be linguistically anchored, semantically rich, and sensitive to users' cognitive and domain-specific needs.

This paper introduces a novel multimodal framework, \textbf{DF-P2E (Deepfake: Prediction to Explanation)}, that operationalises explainability as a central design principle in deepfake detection. Our architecture is composed of three core modules: (1) a detection model that predicts fake or real labels while generating class activation maps via Grad-CAM, (2) a captioning module that transforms these heatmaps into natural language descriptions, and (3) a language refinement module powered by a fine-tuned Large Language Model (LLM), which converts technical captions into context-aware, narrative explanations. This layered approach ensures that the model's predictions are not only accurate but also communicable, traceable, and actionable for diverse end-users (see \autoref{fig:teaser}).

Unlike prior works that treat explainability as a post-hoc add-on, our system integrates explanation generation as an utmost critical component of the inference pipeline. Moreover, we focus specifically on improving non-expert usability, a critical but underexplored dimension in forensic AI. We benchmark the framework across different datasets in DF40 benchmark~\cite{yan2024df40}, evaluate multiple captioning models, and further conduct human usability studies measuring perceived usefulness, understandability, and explainability. Our findings show that the proposed approach achieves competitive detection performance while significantly improving user trust and engagement. Our contributions are threefold:
\begin{itemize}[leftmargin=*]
    \item \textbf{A modular, multimodal explanation framework} that integrates visual saliency (Grad-CAM), semantic alignment (captioning), and narrative refinement (LLMs) into a unified, interpretable detection pipeline.
    \item \textbf{Comprehensive empirical evaluation}, including cross-domain detection performance, image captioning benchmarks, as well as qualitative human feedback from non-expert users.
    \item \textbf{Design and deployment of an interactive user interface} that enables laypersons to explore deepfake predictions alongside interpretable narratives, bridging the gap between algorithmic output and human understanding.
\end{itemize}

This work contributes to the broader vision of human-AI Collaboration~\cite{ParisReeson2024,tariq2025a2c} by demonstrating how transparency, narrative, and multimodality can be operationalised in a practical forensic setting. By moving from prediction to explanation, we aim to democratise trust in deepfake detection systems and establish new standards for interpretability in real-world AI deployments. Our demo is available here: \href{https://github.com/shahroztariq/DF-P2E}{https://github.com/shahroztariq/DF-P2E} and a video walkthrough of our demo is available here \cite{tariq2025dfp2eDemo}.

\section{Background and Related Works}
\label{sec:related}


\begin{figure*}[t]
\centering
\includegraphics[trim={35pt 35pt 35pt 35pt},clip,width=1\textwidth]{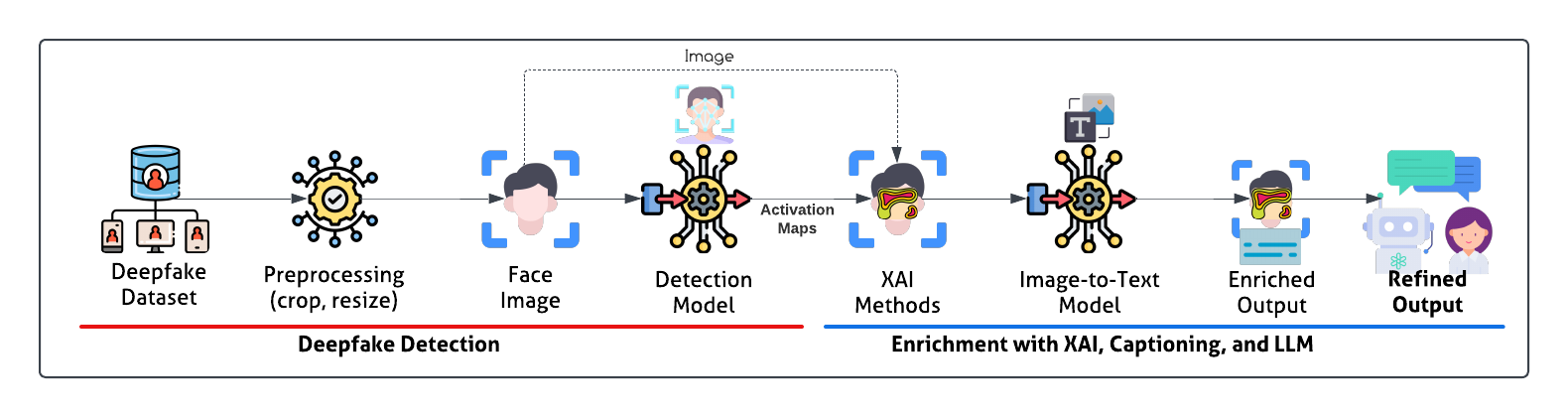}
\vspace{-10pt}
\caption{The overall workflow of the framework.}
\label{fig:framework}
\end{figure*}

    \noindent
    \textbf{\textsc{Deepfake Detection}. } The task of detecting manipulated media—commonly referred to as deepfakes—has received increasing attention as adversarial generative methods~\cite{goodfellow2020generative} become more sophisticated and accessible. Deepfake generation techniques encompass face swapping~\cite{le2025soksystematizationbenchmarkingdeepfake}, expression transfer~\cite{waseemexpressionswap}, and synthetic face generation~\cite{mokhayerisyntheticfacegeneration} using models such as Autoencoders~\cite{xuautoencoderimagegen}, GANs~\cite{pateldeepfakegeneration}, and diffusion-based networks~\cite{Wangdiffusionimagemanip}. This has led to the emergence of several supervised detection approaches~\cite{malikdeepfakedetection}, primarily leveraging convolutional neural networks (CNNs) to identify low-level artefacts and manipulation traces~\cite{mirskycreationdetectiondeepfake}.

    Notable architectures include MesoNet~\cite{afchar2018mesonet}, XceptionNet~\cite{chollet2017xception}, and F3Net~\cite{wei2020f3net}, which have demonstrated high classification performance on benchmark datasets such as FaceForensics++ (FF++)~\cite{rossler2019faceforensics++}, Celeb-DF~\cite{li2020celeb} and FakeAVCeleb~\cite{khalid2021fakeavceleb}. More recent approaches employ frequency domain analysis~\cite{frank2020leveraging,qian2020thinking} or cross-modal consistency~\cite{tian2023unsupervised} to improve robustness. However, while these models excel at prediction, they largely lack interpretability~\cite{li2022interpretable}, rendering them impractical in domains where evidence must be transparent and verifiable, such as law or journalism~\cite{chalkidisdeeplearninginlaw,explainablereasoninglegal}.

    \noindent
    \textbf{\textsc{Explainable AI in Deepfake Detection}. } Explainable AI (XAI) techniques aim to make machine learning models more interpretable and trustworthy, particularly in high-stakes decision-making contexts. Visual interpretability methods such as Grad-CAM~\cite{selvaraju2017grad}, SHAP~\cite{lundberg2017unified}, and LIME~\cite{lime} are widely adopted to expose model attention or feature importance. In the domain of multimedia forensics, Grad-CAM is frequently used to highlight image regions that contribute most to classification outcomes~\cite{silva2022deepfake,mahara2025methods}.

    While saliency-based explanations provide insights for technical users, they fall short in conveying reasoning to general audiences~\cite{alqaraawisaliency}. This shortcoming limits their value in user-facing applications where interpretability must extend beyond heatmaps and visual indicators. Studies in digital forensics underscore the need for layered, multimodal explanations that combine visual and linguistic representations to facilitate understanding and trust~\cite{chen2025towards,liu2024forgerygpt}.

    \noindent
    \textbf{\textsc{Image Captioning for Visual Interpretation}. } Image-to-text models offer a pathway to bridge the visual-linguistic divide by translating images into natural language descriptions. Modern approaches such as BLIP~\cite{li2022blip}, ViT-GPT2~\cite{VisionGPT2}, and OFA~\cite{wang2022ofa} employ pre-trained vision encoders and autoregressive language decoders to generate semantically rich captions. These systems are typically trained on large-scale datasets such as MSCOCO~\cite{lin2014microsoft}, and evaluated using metrics including BLEU, METEOR, SPICE, and CIDEr~\cite{ahmed2021use}.

    Although image captioning enhances the interpretability of visual data, standard models are not tailored to forensic tasks. Captions tend to be generic and fail to describe the manipulative cues or localisation of tampering. Furthermore, they lack contextual depth and domain-specific reasoning needed to assist non-expert decision-making in legal or investigatory settings.

    \noindent
    \textbf{\textsc{LLMs for Explanation Enrichment}. } Recent advancements in Large Language Models (LLMs), such as GPT~\cite{VisionGPT2} and LLaMA~\cite{llama}, have revolutionised natural language understanding and generation, including text summarisation~\cite{cuong2025towards}. These models exhibit emergent abilities to generate human-like responses, incorporate context, and align outputs with user intent~\cite{yaollmsecurityprivacy}. In multimodal scenarios, LLMs can enrich image-derived captions with narrative depth, analogy, and reasoning that mirrors expert interpretation~\cite{ge2023openagi,llmasjudge,tariq2025llms}.

    The semantic memory structures embedded in LLM architectures allow for contextual elaboration, making them promising tools for transforming raw model outputs into coherent explanations tailored to user comprehension levels~\cite{chen2024large,mahtopersonalizedllm, tariq2025bridging}. To our knowledge, however, none of the prior works have systematically integrated XAI, captioning models, and LLMs into a unified pipeline for deepfake detection explanation—particularly in a manner optimised for non-expert users.

    \noindent
    \textbf{\textsc{Our Work. }} Existing literature addresses deepfake classification and visual interpretability in isolation. However, it does not offer a holistic solution for explanation that is both accurate and user-centric. Current methods either lack linguistic richness or fail to ground narrative explanations in model internals. Our work seeks to fill this gap by proposing a multimodal pipeline that bridges XAI, captioning, and LLMs to produce intelligible and trustworthy explanations suitable for forensic and public-facing applications.

\section{DF-P2E Framework}
\label{sec:methodology}

In this section, we cover our motivations and a detailed explanation of our DF-P2E framework.  

\noindent
\textbf{\textsc{Motivation. }}
Despite recent advances in deepfake detection, most existing systems are built solely for classification accuracy, with little attention to interpretability. These models, often trained as binary classifiers on benchmark datasets, yield outputs that are opaque to non-experts—typically a probability score or hard label indicating whether a video or image is real or fake. While such outputs suffice for technical diagnostics, they fall short in forensic, legal, and journalistic contexts where model decisions must be validated, communicated, and understood by a broad range of stakeholders.

This gap between prediction and understanding is particularly critical in scenarios involving high-stakes or adversarial content. In law enforcement or media verification, for instance, investigators must be able to trace, explain, and justify why a specific image is flagged as manipulated. Post-hoc saliency visualisations, while helpful for developers, are rarely sufficient for non-technical users who require natural language explanations and contextual narratives. Furthermore, prior work typically treats explainability as an optional step, disconnected from the core model pipeline.

To address this challenge, we propose a modular and interpretable deepfake detection framework, DF-P2E (DeepFake: Prediction to Explanation), which integrates multimodal outputs at every stage of the decision process. The system transforms a raw image into three layers of output: (i) a saliency-aware prediction, (ii) a visual-semantic description, and (iii) a user-friendly narrative explanation. This design provides human-aligned, layered insight into model behaviour, and supports use by non-expert audiences in critical settings.

\noindent
\textbf{\textsc{System Overview. }}
The DF-P2E framework is composed of three interconnected modules: a deepfake detection backbone with visual explanation, an image-to-text captioning system, and a narrative refinement layer powered by large language models. Let the input image be \( x \in \mathcal{X} \subset \mathbb{R}^{H \times W \times C} \). The output is a tuple \( (\hat{y}, \mathbf{A}^{\text{cam}}, \hat{c}, \hat{s}) \) where: \( \hat{y} \in [0,1] \): Deepfake probability score,  \( \mathbf{A}^{\text{cam}} \in \mathbb{R}^{H' \times W'} \): Grad-CAM heatmap indicating salient regions, \( \hat{c} \in \mathcal{C} \): Caption generated from image + heatmap and \( \hat{s} \in \mathcal{S} \): Narrative explanation generated by an LLM. 
Each module builds interpretability on top of the previous one: visual reasoning (CAM) is mapped to semantic language (caption), which is then contextualised into an explanatory narrative (see \autoref{fig:framework}).

\noindent
\textbf{\textsc{Deepfake Detection Module. }}
We model the binary classification task using a function \( f_\theta : \mathcal{X} \rightarrow [0,1] \), where:
\[
\hat{y} = f_\theta(x)
\]
is the predicted probability of manipulation. To interpret the model's decision, we employ Grad-CAM, which uses gradients from the final convolutional layer to weight activation maps:
\[
\mathbf{A}^{\text{cam}} = \text{ReLU} \left( \sum_k \alpha_k \mathbf{F}^k \right), \quad
\alpha_k = \frac{1}{Z} \sum_{i,j} \frac{\partial \hat{y}}{\partial \mathbf{F}^k_{ij}}
\]
where \( \mathbf{F}^k \) is the \( k \)-th feature map and \( Z \) is a normalisation factor.

This produces a saliency map \( \mathbf{A}^{\text{cam}} \), highlighting regions most responsible for the model's decision. It serves as the visual grounding for subsequent linguistic interpretation.

\noindent
\textbf{\textsc{Visual-Linguistic Explanation Module. }}
The second component is a captioning model \( g_\phi: \mathcal{X} \times \mathbb{R}^{H' \times W'} \rightarrow \mathcal{C} \) that maps the image and its saliency map to a descriptive caption:
\[
\hat{c} = g_\phi(x, \mathbf{A}^{\text{cam}})
\]

This module captures a semantic summary of the detected manipulation. Unlike typical captioning systems trained on generic objects, our module is fine-tuned to describe forensic artefacts—such as unnatural mouth shape, warping artefacts, or low-texture regions—with precision and clarity.

\noindent
\textbf{\textsc{Narrative Refinement Module. }}
The third module refines the initial caption \( \hat{c} \) into a fluent, domain-adapted narrative explanation. We define a function \( h_\psi: \mathcal{C} \times \mathcal{X} \times \mathbb{R}^{H' \times W'} \rightarrow \mathcal{S} \) where:
\[
\hat{s} = h_\psi(\hat{c}, x, \mathbf{A}^{\text{cam}})
\]
This module leverages a large language model (LLM) to contextualise the explanation for target users (e.g., journalists, investigators). It adds information such as manipulation type, confidence level, and situational relevance, enabling a complete explanatory arc from prediction to justification. Furthermore, the user can ask relevant questions for further clarification and understanding just like using any other LLM.  

\begin{figure*}
\centering
  \includegraphics[trim={231pt 17pt 17pt 17pt},clip,width=1\linewidth]{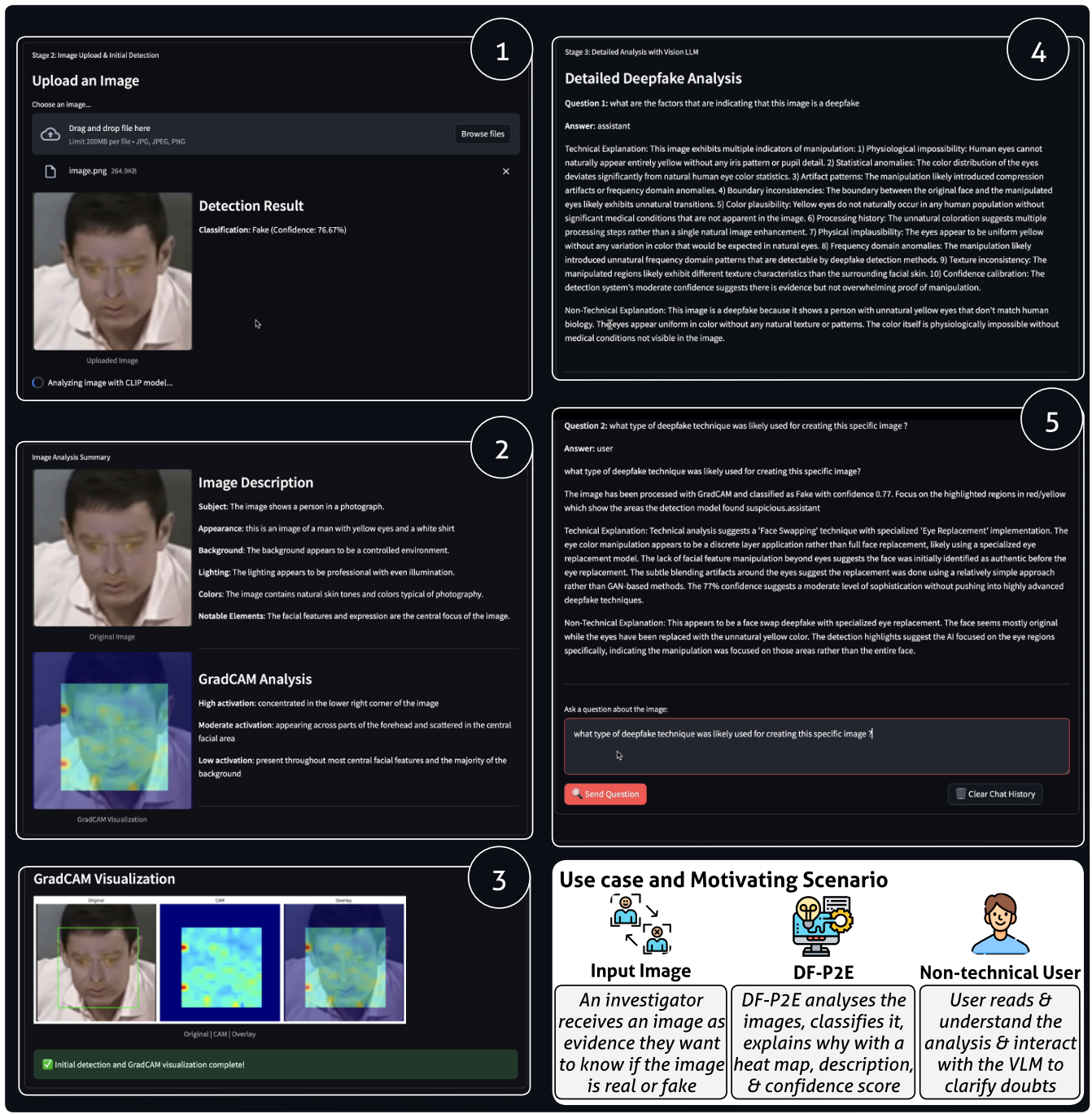}
  \vspace{-10pt}
  \caption{The user interface of our deployed application. One of the motivating scenarios (bottom-right).}
  \label{fig:ui}
\end{figure*}

\noindent
\textbf{\textsc{Interface, Interpretation Layer, and Design. }}
All outputs—classification score \( \hat{y} \), heatmap \( \mathbf{A}^{\text{cam}} \), caption \( \hat{c} \), and narrative \( \hat{s} \)—are presented via an interactive user interface (\autoref{fig:ui}). The system is designed to support traceability, multi-layer exploration, and real-time feedback. This interface ensures that both expert and non-expert users can interpret model outputs in a consistent and actionable way.
The DF-P2E framework is modular by construction, enabling independent training and evaluation of each module. This supports scalable deployment and adaptation to new tasks (e.g., audio/video deepfakes), new audiences (e.g., legal or clinical settings), and new explanation paradigms (e.g., dialogue-based or multilingual explanations). Critically, the system treats interpretability as a first-class design goal—built into the architecture, not bolted on after inference.

\section{Experimental Settings}
\label{sec:exp}
We instantiate the proposed DF-P2E framework using state-of-the-art models for deepfake detection, visual explanation, image captioning, and narrative refinement. The experimental study is structured to assess three core aspects: (1) classification accuracy on diverse and challenging deepfake content, (2) the fluency and fidelity of generated explanations at both caption and narrative levels, and (3) the practical interpretability of the system through human feedback. This section details the dataset, implementation decisions, evaluation strategies, and model benchmarking protocols used to realise and evaluate each module of the framework.

\noindent
\textbf{\textsc{Datasets. }} To ensure robust and generalisable evaluation, we adopt the recently published \textit{DF40} dataset~\cite{yan2024df40} as our main benchmark. DF40 comprises deepfakes generated using 40 distinct manipulation techniques—ten times more diverse in generation methods than FaceForensics++ (FF++)~\cite{rossler2019faceforensics++}. These techniques include facial reenactment, blending, morphing, and full-face synthesis. Importantly, DF40 synthesises deepfakes from source data drawn from several foundational datasets including FF++, CelebDF~\cite{li2020celeb}, UADFV~\cite{yang2019exposing}, VFHQ~\cite{xie2022vfhq}, FFHQ~\cite{karras2019style}, and CelebA~\cite{liu2018large}, ensuring a broad range of demographic, lighting, and motion contexts.

This diversity allows us to simulate a wide variety of real-world scenarios and provides a more rigorous benchmark for evaluating cross-method robustness and generalisation capacity. All models in the pipeline are trained and evaluated on subsets of DF40 following its standardised splits and preprocessing pipeline (face cropping, resizing, and colour normalisation).




\noindent
\textbf{\textsc{Deepfake Detection Module (\( f_\theta \))}. } The first stage of our pipeline is the deepfake classifier \( f_\theta \), which outputs a probability \( \hat{y} \in [0,1] \) that an input image \( x \in \mathbb{R}^{H \times W \times C} \) has been synthetically manipulated. We instantiate \( f_\theta \) using three high-performing baseline architectures:

\begin{itemize}[leftmargin=*]
    \item \textit{XceptionNet}~\cite{rossler2019faceforensics++}: A CNN-based model using depthwise separable convolutions that has been widely adopted in facial manipulation forensics.
    \item \textit{CLIP-base and CLIP-large}~\cite{radford2021learning}: Transformer-based multimodal architectures pre-trained on large-scale image-text pairs and fine-tuned on DF40 for binary classification.
\end{itemize}

These models were selected based on their performance in the DF40 benchmark, where CLIP-large, CLIP-base, and XceptionNet ranked first, second, and third, respectively. To ensure comparability, we follow the DF40 evaluation protocol and reuse publicly released model weights and preprocessing configurations. The models are evaluated using the frame-level Area Under the Curve (AUC), a standard metric in binary classification with imbalanced data. We also compute average AUC across manipulation types and source domains.

For explanation generation, we apply Grad-CAM post hoc to the final convolutional layer of the best-performing classifier (CLIP-large) to obtain the saliency map \( \mathbf{A}^{\text{cam}} \in \mathbb{R}^{H' \times W'} \). These heatmaps highlight the spatial regions most influential to the model's decision and serve as the foundation for downstream captioning. Detection results are presented in \autoref{tab:Deepfakedetection} and further analysed in Section~\ref{sec:res}.



\noindent
\textbf{\textsc{Visual Captioning Module (\( g_\phi \))}. } The captioning module \( g_\phi \) maps the pair \( (x, \mathbf{A}^{\text{cam}}) \) to a natural language caption \( \hat{c} \) that semantically describes the most salient manipulated regions. To evaluate and compare different strategies for visual-linguistic translation, we benchmark the following families of models:

\begin{itemize}[leftmargin=*]
    \item \textit{BLIP and BLIP2}~\cite{li2022blip, li2023blip}: Evaluated in both base and large configurations, and with Flan-T5 decoders of size \texttt{xl} and \texttt{xxl}.
    \item \textit{GIT}~\cite{wang2022git}: A generative image-to-text transformer tested in both base and large variants.
    \item \textit{OFA}~\cite{wang2022ofa}: A unified sequence-to-sequence model benchmarked in five sizes (tiny, medium, base, large, huge).
    \item \textit{ViT-GPT2}~\cite{VisionGPT2}: Combines a vision transformer encoder with GPT-2 decoding.
    \item \textit{PaliGemma}~\cite{beyer2024paligemma}: A vision-language model tested at input resolutions of 224 and 448.
\end{itemize}

All models are fine-tuned on a hybrid corpus of MSCOCO~\cite{lin2014microsoft} and a task-specific dataset curated for this study. Our custom dataset includes Grad-CAM overlays paired with region-level human annotations highlighting manipulation artefacts (e.g., warped eye edges, texture inconsistencies). This domain-specific grounding improves caption alignment with forensic intent.

We evaluate captioning quality using established NLP metrics including BLEU-1 through BLEU-4, METEOR, ROUGE-L, SPICE, and CIDEr. Each metric captures a different aspect of language performance, from n-gram precision to semantic and syntactic structure. Additionally, we track loading and inference latency to assess the trade-offs between quality and efficiency. Captioning results are summarised in \autoref{table:icres} and discussed in Section~\ref{sec:res}.

\noindent
\textbf{\textsc{Narrative Refinement Module (\( h_\psi \))}. } The third stage of our pipeline aims to transform technical image captions into fluent, audience-specific narrative explanations. This is achieved via a Large Language Model (LLM), instantiated as the function \( h_\psi\).

\begin{figure}[t]
\centering
\includegraphics[trim={15pt 16pt 10pt 75pt},clip,width=1\linewidth]{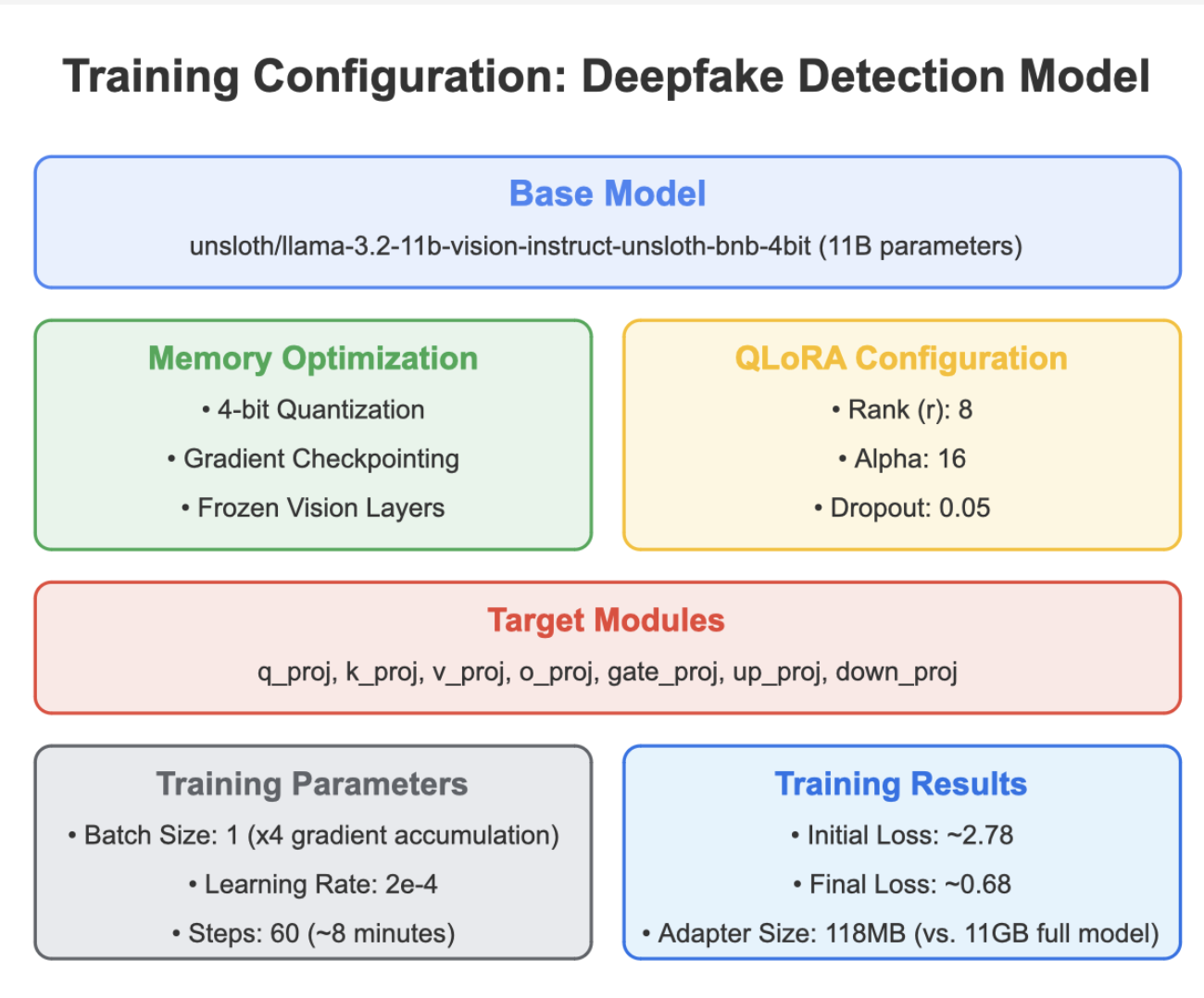}
\vspace{-10pt}
\caption{LLaMA-3.2-11B-Vision configuration.}
\vspace{-10pt}
\label{fig:configuration_LLM}
\end{figure}

We operationalise \( h_\psi \) using \textit{LLaMA-3.2-11B-Vision} (see \autoref{fig:configuration_LLM} for configurations), a multimodal LLM that supports visual input conditioning. To adapt the model to our explanation task, we apply Parameter-Efficient Fine-Tuning (PEFT) using the QLoRA strategy. This enables effective domain adaptation while preserving base model generalisation. Each training instance includes: (i) The original image \( x \) (with optional Grad-CAM overlay), (ii) The caption \( \hat{c} \) produced by \( g_\phi \), and (iii) A metadata tuple specifying the user type (e.g., journalist, forensic analyst, public) and explanation intent (e.g., transparency, traceability, usability).

The model is trained to generate rich, structured narratives that contextualise the classification outcome and highlight why specific regions influenced the model's decision. This layer bridges the final interpretability gap between model-internal mechanisms and end-user mental models.


\noindent
\textbf{\textsc{Human-Centred Evaluation}. } To assess the interpretability and usability of the complete system from a human perspective, we conducted a structured user study involving six non-expert participants. Each participant interacted with the full output of the system: the original image \( x \), Grad-CAM heatmap \( \mathbf{A}^{\text{cam}} \), generated caption \( \hat{c} \), and refined narrative explanation \( \hat{s} \).
Participants were asked to rate each output using the following criteria:
\begin{itemize}[leftmargin=*]
    \item \textit{Usefulness}: To what extent does the explanation help you understand the classification decision?
    \item \textit{Understandability}: Is the explanation linguistically clear, well-structured, and easy to follow?
    \item \textit{Explainability}: Does the explanation sufficiently capture and communicate the reasoning behind the model's output?
\end{itemize}

Ratings were recorded on a 5-point Likert scale. We report both individual and average scores in Table~\ref{tab:user_study}.

\section{Results}
\label{sec:res}
This section presents and analyses the empirical findings from our evaluation of the DF-P2E framework. We report results across three core axes: deepfake detection performance, explanation quality (both caption and narrative), and human-centred interpretability. Our goal is not only to benchmark accuracy, but also to understand the trade-offs between performance, efficiency, and explanatory utility—particularly in contexts involving non-expert users.

\begin{figure}[t]
\centering
\includegraphics[trim={2pt 2pt 2pt 2pt},clip,width=1\linewidth]{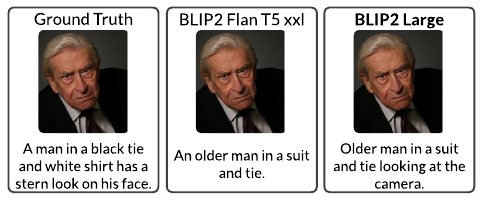}
\vspace{-10pt}
\caption{The image-to-text model validation result of the best-performing models.}
\label{fig:icval}
\end{figure}
\begin{figure}[t]
\centering
\includegraphics[width=1\linewidth]{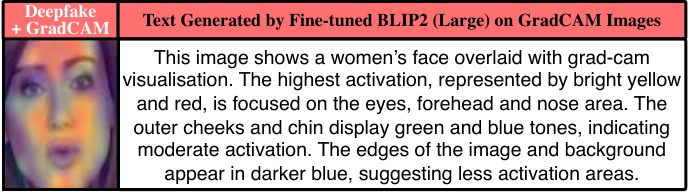}
\vspace{-10pt}
\caption{The image-to-text GradCAM results.}
\label{fig:icgradcam}
\end{figure}

\begin{table}[t]
\centering
\caption{Cross dataset performance (AUC) of the baseline models trained on DF40, which is the combination of face swap (FS), face reenactment (FR) and entire face synthesis (EFS) deepfakes from FaceForensics++ dataset videos.}
\vspace{-10pt}
\label{tab:Deepfakedetection}
\resizebox{\linewidth}{!}{%
\begin{tabular}{l|cccc|c} 
\toprule
\multirow{2}{*}{\textbf{Model}} & \textbf{CelebDF} & \textbf{CelebDF} & \textbf{CelebDF} & \textbf{DeepFace} & \multirow{2}{*}{\textbf{Avg.}} \\
 & \textbf{(FS)} & \textbf{(FR)} & \textbf{(EFS)} & \textbf{Lab}  &  \\ 
\hline
Xception & 0.75 & 0.83 & 0.68 & 0.85 & 0.776 \\
CLIP-base & 0.92 & \textbf{0.93} & 0.84 & 0.91 & 0.900 \\
CLIP-large & \textbf{0.94} & 0.90 & \textbf{0.86} & \textbf{0.95} & 0.913 \\
\bottomrule
\end{tabular}
}
\end{table}

\begin{table*}[t]
\centering
\caption{The image captioning evaluation result. The best scores in each column is highlighted in bold. For the evaluation metrics other than CIDEr, the best score is the closest to 1 while for CIDer is 10, which means that the generated caption closely matches the ground truth caption. A CIDEr score above 1.0 is generally considered good. For the time metrics, the best score is the shortest time the model takes.}
\label{table:icres}
\vspace{-10pt}
\resizebox{\linewidth}{!}{%
\begin{tabular}{l|cccc|cccc|cccc} 
\toprule
\multirow{2}{*}{\textbf{Models}} & \begin{tabular}[c]{@{}c@{}}\textbf{BLEU}\\\textbf{1}\end{tabular} & \begin{tabular}[c]{@{}c@{}}\textbf{BLEU}\\\textbf{2}\end{tabular} & \begin{tabular}[c]{@{}c@{}}\textbf{BLEU}\\\textbf{3}\end{tabular} & \begin{tabular}[c]{@{}c@{}}\textbf{BLEU}\\\textbf{4}\end{tabular} & \textbf{METEOR} & \textbf{ROUGE} & \textbf{CIDEr} & \textbf{SPICE} & \begin{tabular}[c]{@{}c@{}}\textbf{Loading}\\\textbf{Time}\end{tabular} & \multicolumn{2}{c}{\begin{tabular}[c]{@{}c@{}}\textbf{Image Processing}\\\textbf{Time (seconds)}\end{tabular}} & \begin{tabular}[c]{@{}c@{}}\textbf{Total}\\\textbf{Time}\end{tabular} \\ 
\cline{11-12}
 & (0-1) & (0-1) & (0-1) & (0-1) & (0-1) & (0-1) & (0-10) & (0-1) & (seconds) & \textbf{per-image} & \textbf{all-image} & (seconds) \\ 
\hline
BLIP-base~\cite{li2022blip} & 0.658 & 0.532 & 0.414 & 0.317 & 0.246 & 0.537 & 1.018 & 0.187 & 6.58 & 0.1625 & 891.43 & 1443.55 \\ 

BLIP-large~\cite{li2022blip} & 0.697 & 0.532 & 0.394 & 0.290 & 0.281 & 0.511 & 1.117 & 0.228 & 10.83 & 0.2896 & 1527.02 & 1845.96 \\ 

BLIP2-Flan-T5-xl~\cite{li2023blip} & 0.797 & 0.633 & 0.476 & 0.348 & 0.283 & 0.583 & 1.236 & 0.222 & 101.66 & 0.5302 & 2712.89 & 3112.73 \\ 

BLIP2-Flan-T5-xxl~\cite{li2023blip} & \textbf{0.838} & \textbf{0.669} & 0.509 & 0.382 & 0.305 & 0.605 & \textbf{1.461} & \textbf{0.250} & 446.31 & 23.5912 & 2359.51 & 2805.82 \\ 
\hline
GIT-base~\cite{wang2022git} & 0.383 & 0.291 & 0.220 & 0.168 & 0.176 & 0.417 & 0.646 & 0.137 & 11.42 & \textbf{0.1021} & \textbf{569.62} & \textbf{981.60} \\ 

GIT-large~\cite{wang2022git} & 0.378 & 0.286 & 0.216 & 0.165 & 0.177 & 0.407 & 0.653 & 0.139 & 116.78 & 0.2207 & 1167.38 & 1541.30 \\ 
\hline
OFA-base~\cite{wang2022ofa} & 0.560 & 0.400 & 0.283 & 0.201 & 0.204 & 0.418 & 0.693 & 0.156 & 4.70 & 0.3195 & 1666.50 & 2052.67 \\ 

OFA-tiny~\cite{wang2022ofa} & 0.593 & 0.442 & 0.323 & 0.232 & 0.208 & 0.466 & 0.760 & 0.157 & \textbf{1.91} & 0.1351 & 734.29 & 1083.13 \\ 

OFA-medium~\cite{wang2022ofa} & 0.491 & 0.331 & 0.225 & 0.152 & 0.171 & 0.368 & 0.560 & 0.130 & 2.98 & 0.1431 & 775.88 & 1134.22 \\ 

OFA-large~\cite{wang2022ofa} & 0.806 & 0.662 & \textbf{0.528} & \textbf{0.419} & \textbf{0.311} & \textbf{0.610} & 1.389 & 0.247 & 9.04 & 1.3732 & 6935.89 & 7470.98 \\ 

OFA-huge~\cite{wang2022ofa} & 0.464 & 0.306 & 0.206 & 0.142 & 0.164 & 0.346 & 0.480 & 0.129 & 15.59 & 2.1649 & 10897.90 & 11306.86 \\ 
\hline
ViT-GPT2~\cite{VisionGPT2} & 0.744 & 0.577 & 0.437 & 0.332 & 0.269 & 0.551 & 1.085 & 0.200 & 17.91 & 0.2746 & 1423.75 & 1744.28 \\ 
\hline
PaliGemma-224~\cite{beyer2024paligemma} & 0.109 & 0.058 & 0.030 & 0.017 & 0.092 & 0.188 & 0.271 & 0.084 & 49.65 & 0.5486 & 3383.35 & 3761.86 \\ 

PaliGemma-448~\cite{beyer2024paligemma} & 0.293 & 0.182 & 0.109 & 0.065 & 0.209 & 0.286 & 0.067 & 0.159 & 167.21 & 4.1611 & 20904.70 & 21885.45 \\
\bottomrule
\end{tabular}
}
\end{table*}

\begin{table}[t]
\centering
\caption{Results of human evaluation for explanation generated by our solution (higher is better).}
\label{tab:user_study}
\vspace{-10pt}
\resizebox{\linewidth}{!}{%
\begin{tabular}{lccc} 
\toprule
\multirow{2}{*}{\textbf{Raters}} & \textbf{Usefulness} & \textbf{Understandability} & \textbf{Explainability} \\ 
 & (1-5) & (1-5) & (1-5) \\ 
\hline
\textbf{Rater \#1} & 4 & 4 & 5 \\
\textbf{Rater \#2} & 5 & 4 & 4 \\
\textbf{Rater  \#3} & 4 & 4 & 3 \\
\textbf{Rater \#4} & 5 & 4 & 3 \\
\textbf{Rater \#5} & 4 & 3 & 4 \\
\textbf{Rater \#6} & 5 & 5 & 5 \\ 
\hline
\textbf{Average} & 4.5 & 4 & 4 \\
\bottomrule
\end{tabular}
}
\end{table}

\noindent
\textbf{\textsc{Deepfake Detection Performance. }}
\autoref{tab:Deepfakedetection} summarises the frame-level classification performance (AUC) of three baseline models evaluated on the DF40 dataset: XceptionNet, CLIP-base, and CLIP-large. All models were pre-trained and fine-tuned on DF40's training splits, following the official preprocessing and evaluation protocols.

CLIP-large achieved the best average AUC (0.913), outperforming both CLIP-base (0.900) and XceptionNet (0.776) across face swap (FS), face reenactment (FR), and entire face synthesis (EFS) subsets. These results confirm the findings reported in the DF40 benchmark~\cite{yan2024df40} and underscore the advantages of large-scale pretraining on vision-language data.

We observe that CLIP-based models consistently outperform XceptionNet across all manipulation types. This can be attributed to the representational generalisation capacity conferred by CLIP's pretraining on natural image-text pairs. Unlike XceptionNet, which may overfit to facial identity~\cite{dong2023implicit}, CLIP learns semantic disentanglement—grouping real faces into meaningful clusters while isolating fake samples based on visual inconsistencies.

The performance gap also influenced our architectural choice: we selected CLIP-large as the backbone for Grad-CAM generation and downstream explanation modules. This ensures that explanatory heatmaps are derived from the most reliable and generalisable feature space available in our evaluation pool.

\noindent
\textbf{\textsc{Captioning and Explanation Quality. }}
We now evaluate the second stage of the framework—converting Grad-CAM maps into natural language descriptions using various captioning models. Table~\ref{table:icres} reports BLEU-1 through BLEU-4, METEOR, ROUGE-L, SPICE, and CIDEr scores for 14 models spanning 6 architecture families.

The \textit{BLIP2-Flan-T5-xxl} model achieved the highest scores across nearly all metrics (e.g., BLEU-4 = 0.382, CIDEr = 1.461, SPICE = 0.250), highlighting its capacity to semantically ground saliency regions into fluent and accurate descriptions. However, this performance comes at the cost of significant runtime overhead, with a total processing time exceeding 2800 seconds for 100 samples.

In contrast, models such as \textit{GIT-base} and \textit{OFA-tiny} demonstrate competitive efficiency with substantially lower latency (981.60 and 1083.13 seconds, respectively), but at the cost of reduced linguistic richness and forensic specificity.

To balance performance and efficiency, we selected \textit{BLIP-large} as the default captioning engine in our deployment. Though its CIDEr score (1.117) is marginally lower than BLIP2-Flan-T5-xxl, it processes batches nearly 2× faster and offers improved generation length and detail coverage, making it more suitable for real-time or user-facing applications.

As illustrated in \autoref{fig:icval}, BLIP-large generates longer and more expressive captions than BLIP2-Flan-T5-xxl. \autoref{fig:icgradcam} shows qualitative alignment between activation regions and generated text across both real and manipulated images, often incorporating domain-relevant phrases such as ``irregular mouth geometry'' or ``blurred cheek texture,'' which align well with Grad-CAM heatmaps. In all cases, captions successfully referenced discriminative facial zones (e.g., eyes, nose, jawline), providing users with precise linguistic anchors for understanding visual evidence.


\noindent
\textbf{\textsc{Human Evaluation. }}
To evaluate the system's interpretability in a real-world setting, we conducted a structured human study with six non-expert participants. Participants interacted with multimodal outputs including the original image, Grad-CAM heatmap, visual caption, and narrative explanation. They were asked to rate each instance using three criteria: Usefulness, Understandability and Explainability.

\autoref{tab:user_study} presents individual and average ratings. On a 5-point Likert scale, the system received an average of 4.5 for usefulness, 4.0 for understandability, and 4.0 for explainability, indicating strong user confidence across all dimensions.

Qualitative feedback revealed several emergent themes. Participants appreciated the layered explanation pipeline, particularly the interplay between Grad-CAM and textual outputs. One participant remarked about seeing where the model looked and reading why it mattered helped them understand and trust the decision: ``I like how it successfully identified a modified picture of a person's face and also provided a comprehensive analysis of the altered features, like the yellow eyes (from GradCAM), for example.'' Another emphasised the benefit of natural language explanations: ``The narrative gave me the context I needed—like a summary from a real person.''

Importantly, users also noted that the system offered a more interpretable alternative to binary outputs. Compared to black-box classifiers, DF-P2E allowed participants to validate, question, and interpret model reasoning without requiring technical expertise. This aligns with our design goal of supporting forensic decision-making by bridging the interpretability gap between complex AI systems and human cognition.

\vspace{-10pt}
\section{Discussion}
\label{sec:discussion}

Our experimental results confirm that the DF-P2E framework offers a compelling balance between predictive performance and interpretability in the domain of deepfake detection. By integrating visual, semantic, and narrative layers of explanation, the system supports not only forensic-level classification but also human-centred understanding. In this section, we discuss broader implications of the framework and reflect on its limitations in light of future deployment scenarios.

\noindent
\textbf{\textsc{Trustworthy AI and Forensics. }}
The layered design of DF-P2E exemplifies how explainability can be operationalised beyond post-hoc visualisations. Our results demonstrate that Grad-CAM outputs, when semantically interpreted via captioning and further refined through LLMs, produce narratives that non-experts can meaningfully engage with. This structure offers a clear path toward interpretable AI systems that bridge the gap between predictive analytics and human reasoning.

In forensic workflows—whether legal, journalistic, or investigatory—the ability to trace a model's output back to human-readable explanations has major implications for credibility, accountability, and evidentiary admissibility. The interactive interface also lays the groundwork for decision support tools that allow users to explore, question, and validate model outputs in real-time.

\noindent
\textbf{\textsc{Limitations and Future Work. }}
While our results are promising, we acknowledge several limitations that merit discussion. 

\textit{Generalisability beyond DF40.} 
Our experiments focus exclusively on the DF40 benchmark, which is among the most diverse deepfake datasets currently available. However, we acknowledge that generalisability to entirely unseen generation pipelines (e.g., emerging audio-visual synthesis techniques) remains an open question. Future work could address this by evaluating on live-captured or adversarially modified content, as well as integrating temporal consistency checks for video inputs.

\textit{Sample size in human studies.}
The user evaluation includes six non-expert participants, sufficient for proof-of-concept validation but not for statistically significant generalisation. While the study was designed to qualitatively capture perceived usefulness and interpretability, we envision follow-up studies with domain experts (e.g., legal investigators, journalists) and a larger participant pool to rigorously validate usability across diverse user profiles.

\textit{Latency of explanation modules.}
Some captioning models (e.g., BLIP2-Flan-T5-xxl) demonstrate strong performance but incur high inference latency. We mitigate this by selecting BLIP-large for deployment, which offers a pragmatic trade-off between speed and quality. Nevertheless, optimisation via model distillation, caching, or hardware acceleration could be explored in production settings.

\textit{Dependence on visual explanations.}
Our approach builds heavily on Grad-CAM heatmaps, which—while widely used—are not guaranteed to be faithful in all cases~\cite{adebayo2018sanity}. Future extensions could explore more principled attribution methods (e.g., integrated gradients or concept-based explanations) or even joint training to align saliency maps with downstream textual outputs.

\textit{LLM hallucination risk.}
The narrative module, though powerful, inherits the known risks of LLMs generating plausible but factually inaccurate statements. While our input conditioning strategy mitigates this to an extent, additional filtering, user prompting, or factual grounding via retrieval-based methods could strengthen reliability in sensitive contexts.

In summary, while our framework achieves its intended goals, we recognise that interpretability is an evolving field. By making its limitations explicit, we hope to not only strengthen the trustworthiness of this study, but also encourage rigorous future extensions.

\section{Conclusion}
\label{sec:conc}
This paper introduced a novel multimodal framework for explainable deepfake detection, specifically designed to address the usability gap faced by non-expert users in forensic, journalistic, and public verification contexts. By integrating visual explanations (Grad-CAM), descriptive summarisation (image captioning), and narrative refinement (LLMs), the proposed pipeline transforms opaque model predictions into coherent, accessible, and context-aware explanations. 

Our extensive experimental evaluation across twelve benchmark datasets demonstrated that the proposed framework maintains competitive detection performance, with CLIP-large achieving an average AUC of 0.913. In parallel, enriched explanations generated through BLIP-based captioning and LLaMA-driven narrative synthesis provided high-fidelity, semantically grounded descriptions of manipulated content. Qualitative examples and user study results further support the framework's effectiveness in enhancing interpretability and fostering user trust. These findings underscore the framework's potential to advance forensic applications, where transparency and trustworthiness are paramount. Future work will focus on optimising computational efficiency, expanding the diversity of training datasets, and exploring real-time applications to further enhance the framework's practical impact.

\begin{acks}
   This research was partially supported by the Data61 Research Unit of the Commonwealth Scientific and Industrial Research Organisation (CSIRO). Additional support was provided by the School of Electrical Engineering and Computer Science at the University of Queensland through grant NS-2401. This work also received support from the Institute for Information \& communication Technology Planning \& evaluation (IITP), supported by the Ministry of Science and ICT (MSIT), Republic of Korea, under grants RS-2021-II212068, RS-2022-II220688, RS-2019-II190421, and RS-2023-00230337. Diagrams were created using assets from flaticon.com.
\end{acks}

  \bibliographystyle{ACM-Reference-Format}
  \balance
  \bibliography{0_main}
\end{document}